\documentclass[conference]{IEEEtran}

\usepackage{hyperref}
\usepackage{graphicx}
\usepackage{textcomp}
\usepackage{xcolor}
\usepackage[export]{adjustbox}
\usepackage{amsthm}
\usepackage{amsmath}
\usepackage{amssymb}
\usepackage{mathtools}
\usepackage{multirow}
\usepackage{pgf}
\usepackage{tikz}
\usetikzlibrary{arrows, automata, shapes, petri, positioning, calc}
\usepackage{caption}
\usepackage{pgfplots}
\usepackage[export]{adjustbox}

\hypersetup{
	colorlinks=true,
	linkcolor={red!50!black},
	citecolor={blue!60!black},
	urlcolor={blue!80!black},
	pdfauthor={Pegoraro Marco},
	pdftitle={{Process Mining on Uncertain Event Data}},
	pdfsubject={Process Mining over Uncertain Data},
	pdfkeywords={Process Mining, Uncertain Data, Partial Order},
	pdfproducer={LaTeX},
	pdfcreator={pdfLaTeX},
	bookmarksopen=true
}

\newtheorem{assertion}{Assertion}

\begin{document}

\title{Process Mining on Uncertain Event Data\\(Extended Abstract)}

\author{\IEEEauthorblockN{Marco Pegoraro*}
	\IEEEauthorblockA{Chair of Process and Data Science (PADS)\\
		Department of Computer Science, RWTH Aachen University, Ahornstr. 55, 52074 Aachen, Germany\\
		\,*Corresponding author. Email: \url{pegoraro@pads.rwth-aachen.de}}
}

\maketitle

\begin{abstract}
With the widespread adoption of process mining in organizations, the field of process science is seeing an increase in the demand for ad-hoc analysis techniques of non-standard event data. An example of such data are \emph{uncertain event data}: events characterized by a described and quantified attribute imprecision. This paper outlines a research project aimed at developing process mining techniques able to extract insights from uncertain data. We set the basis for this research topic, recapitulate the available literature, and define a future outlook.
\end{abstract}

\IEEEpeerreviewmaketitle

\section{Introduction}\label{sec:introduction}
Since its inception, process mining has ultimately proved its value in commercial applications. An ever-increasing number of success stories has led to a vast demand of the most diverse process analysis techniques, often customized to meet the needs of specific domains. Among these, novel techniques have been introduced to mine non-standard types of data.

This paper presents a research direction aimed to mine one such type of anomalous (i.e, uncommon) type of event data: \emph{uncertain data}. Such data is associated with a degree of imprecision that affects event attributes, which is described and quantified through sets of possible attribute labels, intervals of possible values, or probability distributions.

The remainder of the paper is structured as follows.
Section~\ref{sec:data} illustrates with examples the structure of uncertain event data.
Section~\ref{sec:approach} shows the research principles in regard of process mining on uncertain data, and reports recent results on the topic. Finally, Section~\ref{sec:challenges} outlines open challenges, outlook, and future perspectives of this line of research.

\section{Uncertain Data}\label{sec:data}
In order to more clearly visualize the structure of the attributes in uncertain events, let us consider the following process instance, which is a simplified version of actually occurring anomalies, e.g., in the processes of the healthcare domain. An elderly patient enrolls in a clinical trial for an experimental treatment against myeloproliferative neoplasms, a class of blood cancers. This enrollment includes a lab exam and a visit with a specialist; then, the treatment can begin. The lab exam, performed on the 8th of July, finds a low level of platelets in the blood of the patient, a condition known as thrombocytopenia (TP). During the visit on the 10th of July, the patient reports an episode of night sweats on the night of the 5th of July, prior to the lab exam. The medic notes this but also hypothesizes that it might not be a symptom, since it can be caused either by the condition or by external factors (such as very warm weather). The medic also reads the medical records of the patient and sees that, shortly prior to the lab exam, the patient was undergoing a heparin treatment (a blood-thinning medication) to prevent blood clots. The thrombocytopenia, detected by the lab exam, can then be either primary (caused by the blood cancer) or secondary (caused by other factors, such as a concomitant condition). Finally, the medic finds an enlargement of the spleen in the patient (splenomegaly). It is unclear when this condition has developed: it might have appeared at any moment prior to that point. These events are recorded in the trace ID192-1 (shown in Table~\ref{table:uncertaintraces}) within the hospital's information system.

Such scenario, with no known probability, is known as \emph{strong uncertainty}. In this trace, the rightmost column refers to event indeterminacy: in this case, $e_1$ has been recorded, but it might not have occurred in reality, and is marked with a ``?'' symbol. Event $e_2$ has more then one possible activity labels, either \emph{PrTP} or \emph{SecTP}. Lastly, event $e_3$ has an uncertain timestamp, and might have happened at any point in time between the 4th and 10th of July.

Uncertain events may also have probability values associated with them, a scenario defined as \emph{weak uncertainty} (trace ID192-2 in Table~\ref{table:uncertaintraces}). In the example described above, suppose the medic estimates that there is a high chance (90\%) that the thrombocytopenia is primary (caused by the cancer). Furthermore, if the splenomegaly is suspected to have developed three days prior to the visit, which takes place on the 10th of July, the timestamp of event $e_3$ may be described through a Gaussian curve with $\mu = 7$. Lastly, the probability that the event $e_1$ has been recorded but did not occur in reality may be known (for example, it may be 25\%).

\begin{table}[h]
	\caption{Two uncertain traces related to an example of healthcare process. The timestamps column shows only the day of the month.}
	\label{table:uncertaintraces}
	\centering
	\begin{tabular}{ccccc}
		\textbf{Case ID}        & \textbf{Event ID} & \textbf{Timestamp}                                                                                                     & \textbf{Activity}             & \multicolumn{1}{l}{\textbf{Indeterminacy}} \\ \hline
		\multicolumn{1}{|c|}{ID192-1} & \multicolumn{1}{c|}{$e_1$} 
		& \multicolumn{1}{c|}{5}                                                                         & \multicolumn{1}{c|}{\emph{NightSweats}}        & \multicolumn{1}{c|}{?}                    \\ \hline
		\multicolumn{1}{|c|}{ID192-1}& \multicolumn{1}{c|}{$e_2$} & \multicolumn{1}{c|}{8}                                                                         & \multicolumn{1}{c|}{\emph{PrTP}, \emph{SecTP}} & \multicolumn{1}{c|}{}                    \\ \hline
		\multicolumn{1}{|c|}{ID192-1}& \multicolumn{1}{c|}{$e_3$} & \multicolumn{1}{c|}{4--10}                                                                         & \multicolumn{1}{c|}{\emph{Splenomeg}} & \multicolumn{1}{c|}{}                    \\ \hline
		\multicolumn{1}{|c|}{ID192-2} & \multicolumn{1}{c|}{$e_4$} 
		& \multicolumn{1}{c|}{5}                                                                         & \multicolumn{1}{c|}{\emph{NightSweats}}        & \multicolumn{1}{c|}{$?: 25\%$}                    \\ \hline
		\multicolumn{1}{|c|}{ID192-2}& \multicolumn{1}{c|}{$e_5$} & \multicolumn{1}{c|}{8}                                                                         & \multicolumn{1}{c|}{\begin{tabular}[c]{@{}c@{}}\emph{PrTP: $90\%$},\\ \emph{SecTP: $10\%$}\end{tabular}} & \multicolumn{1}{c|}{}                    \\ \hline
		\multicolumn{1}{|c|}{ID192-2}& \multicolumn{1}{c|}{$e_6$} & \multicolumn{1}{c|}{$\mathcal{N}(7, 1)$}                                                                         & \multicolumn{1}{c|}{\emph{Splenomeg}} & \multicolumn{1}{c|}{}                    \\ \hline
	\end{tabular}
\end{table}


Table~\ref{table:unctypes} summarizes the types of uncertain data subject of our research.

\pgfmathdeclarefunction{gauss}{2}{%
	\pgfmathparse{1/(#2*sqrt(2*pi))*exp(-((x-#1)^2)/(2*#2^2))}%
}

\begin{table}[h]
	\centering
	\begin{adjustbox}{width=.98\linewidth,center}
		\begin{tabular}{|c|c|c|}
			\hline
			& \textbf{Weak uncertainty} & \textbf{Strong uncertainty} \\
			& \textbf{(stochastic)} & \textbf{(non-deterministic)} \\ \hline
			\textbf{Discrete data}   &
			\begin{tabular}[c]{@{}c@{}} Discrete probability distribution\\ \\
				\begin{minipage}{.15\textwidth}
					\resizebox {\textwidth} {!} {
						\begin{tikzpicture}
						\begin{axis}[ybar interval, ymax=55,ymin=0, minor y tick num = 3]
						\addplot coordinates { (0, 5) (5, 35) (10, 50) (15, 30) (20, 15) (25, 0) };
						\end{axis}
						\end{tikzpicture}
					}
				\end{minipage}
			\end{tabular} &
			\begin{tabular}[c]{@{}c@{}} Set of possible values \\ \\ $\{ x_1, x_2, x_3, \dots \} \subseteq X $
			\end{tabular} \\ \hline
			\textbf{Continuous data} &
			\begin{tabular}[c]{@{}c@{}} Probability density function \\ \\
				\begin{minipage}{.15\textwidth}
					\resizebox {\textwidth} {!} {
						\begin{tikzpicture}
						\begin{axis}[every axis plot post/.append style={
							mark=none,domain=-2:3,samples=50,smooth}, 
						axis x line*=bottom, 
						axis y line*=left, 
						enlargelimits=upper]
						\addplot {gauss(1,0.5)};
						\end{axis}
						\end{tikzpicture}
					}
				\end{minipage}
			\end{tabular} &
			\begin{tabular}[c]{@{}c@{}} Interval \\ \\ $\{ x \in \mathbb{R} \mid a \leq x \leq b \}$
			\end{tabular} \\ \hline
		\end{tabular}
	\end{adjustbox}
	\caption{The four different types of uncertainty subject of this research project.}
	\label{table:unctypes}
\end{table}

\section{Research Approach}\label{sec:approach}
We will now illustrate the guiding principles of our research plans, through a series of assertions.

\begin{assertion}[Uncertainty is not noise]
	Uncertain data contain information and value. We do not aim to analyze the data beyond the uncertainty, but the data within the uncertainty.
\end{assertion}

\begin{assertion}[Uncertainty should not be filtered or repaired]
	To extract information from uncertainty itself, existing approaches to filter or repair data are not applicable: information from uncertainty must be accounted for, and not altered.
\end{assertion}

\begin{assertion}[Uncertainty is behavior]
	The many possible values for event attributes entail numerous possible scenarios for the control-flow perspective of an uncertain trace---which can be represented as behavior. To fully analyze uncertain process instances, it is necessary to account for such behavior.
\end{assertion}

\begin{figure}[h]
	\centering
	\includegraphics[width=1\columnwidth]{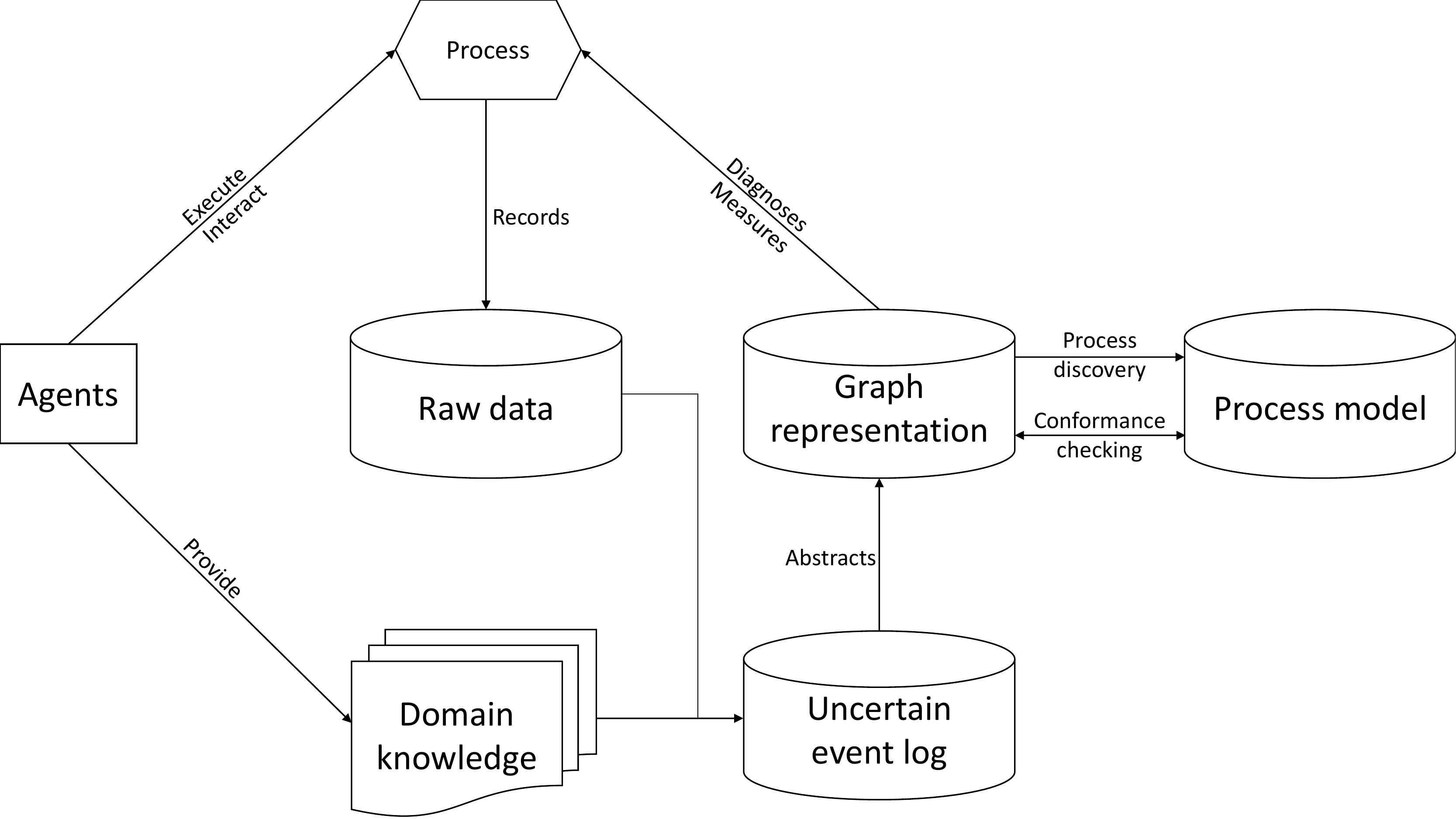}
	\caption{The overall schema for process mining over uncertainty.}
	\label{fig:schema}
\end{figure}

The fundamental technique that enables the analysis of uncertain traces is their representation as dynamic objects that incorporate the intrinsic behavior of uncertain traces, such as graphs or Petri nets (\emph{behavior graphs} or \emph{behavior nets}~\cite{pegoraro2019mining}, respectively). This leads to the schematic visible in Figure~\ref{fig:schema}.

A number of mining techniques for uncertain event data are now present in literature. A taxonomy of uncertain event data is available~\cite{pegoraro2019mining}, as well as a method to reliably compute the probability associated with each real-life scenario in an uncertain trace~\cite{pegoraro2021probability}. There exist approaches for conformance checking~\cite{pegoraro2021conformance} and process discovery~\cite{pegoraro2019discovering} over strongly uncertain event data. The key phase in uncertain data analysis of building graph representation has been optimized through efficient algorithms~\cite{pegoraro2020efficient,pegoraro2020efficient2}. Such techniques are available in the PROVED toolset~\cite{pegoraro2021proved}, which employs an ad-hoc extension of the XES standard to represent uncertain data~\cite{pegoraro2021an}. A real-life source of uncertain data, convolutional neural network sensing in video feeds of processes, has been described, as well as an additional taxonomy also involving process models~\cite{DBLP:journals/corr/abs-2106-03324}.

\section{Open Challenges and Conclusion}\label{sec:challenges}
The field of process mining over uncertain data is still in its infancy. While some techniques to perform discovery and conformance checking over uncertainty do exist, the weakly uncertain case is still unexplored. The principle of the four quality metrics of logs and processes (fitness, precision, simplicity, precision), a cornerstone of process mining, needs to be (re)developed in the context of uncertain data.

Through analyzing uncertain event data without discarding any of the attributes in an uncertain event log, this research direction unlocks the extraction of process information formerly inaccessible. Insights from process mining analyses can, as a consequence, maintain quantified guarantees of reliability and accuracy even in presence of data affected by uncertainty.

\section*{Acknowledgments}
I am very grateful to Prof. Wil van der Aalst, who advises my doctoral studies, and to Merih Seran Uysal, who supervises me in researching this topic. I thank the Alexander von Humboldt (AvH) Stiftung for supporting my research interactions.

\bibliographystyle{IEEEtran}
\bibliography{bibliography.bib}

\end{document}